\documentclass[conference]{IEEEtran}
\IEEEoverridecommandlockouts
\usepackage{cite}
\usepackage{amsmath,amssymb,amsfonts}
\usepackage{algorithmic}
\usepackage{graphicx}
\usepackage{textcomp}
\usepackage{xcolor}
\def\BibTeX{{\rm B\kern-.05em{\sc i\kern-.025em b}\kern-.08em
    T\kern-.1667em\lower.7ex\hbox{E}\kern-.125emX}}

\usepackage{orcidlink}

\makeatletter
\newcommand{\linebreakand}{%
  \end{@IEEEauthorhalign}
  \hfill\mbox{}\par
  \mbox{}\hfill\begin{@IEEEauthorhalign}
}
\makeatother

\usepackage{pgfplots}
\pgfplotsset{compat=1.7}

\begin{document}

\title{LawPal : A Retrieval Augmented Generation Based System for Enhanced Legal Accessibility in India\\

}

\author{\IEEEauthorblockN{Dnyanesh Panchal}
\IEEEauthorblockA{\textit{Bachelor of Engg, Dept. of AI \& DS} \\
\textit{Vidyavardhini's College of Engineering and Technology}\\
MH, India \\
dnyanesh.225017108@vcet.edu.in}
\and
\IEEEauthorblockN{ Aaryan Gole}
\IEEEauthorblockA{\textit{Bachelor of Engg, Dept. of AI \& DS} \\
\textit{Vidyavardhini's College of Engineering and Technology}\\
MH, India \\
aaryan.224767101@vcet.edu.in}
\linebreakand
\IEEEauthorblockN{Vaibhav Narute}
\IEEEauthorblockA{\textit{Bachelor of Engg, Dept. of AI \& DS} \\
\textit{Vidyavardhini's College of Engineering and Technology}\\
MH, India \\
vaibhav.224997106@vcet.edu.in}
\and
\IEEEauthorblockN{Raunak Joshi\,\orcidlink{0009-0009-1796-4213}}
\IEEEauthorblockA{\textit{Assistant Professor, Dept. of AI \& DS} \\
\textit{Vidyavardhini's College of Engineering and Technology}\\
MH, India \\
raunak.joshi@vcet.edu.in}
}

\maketitle

\begin{abstract}
Access to legal knowledge in India is often hindered by a lack of awareness, misinformation and limited accessibility to judicial resources. Many individuals struggle to navigate complex legal frameworks, leading to the frequent misuse of laws and inadequate legal protection. To address these issues, we propose a Retrieval-Augmented Generation (RAG)-based legal chatbot powered by vectorstore oriented FAISS for efficient and accurate legal information retrieval. Unlike traditional chatbots, our model is trained using an extensive dataset comprising legal books, official documentation and the Indian Constitution, ensuring accurate responses to even the most complex or misleading legal queries. The chatbot leverages FAISS for rapid vector-based search, significantly improving retrieval speed and accuracy. It is also prompt-engineered to handle twisted or ambiguous legal questions, reducing the chances of incorrect interpretations. Apart from its core functionality of answering legal queries, the platform includes additional features such as real-time legal news updates, legal blogs, and access to law-related books, making it a comprehensive resource for users. By integrating advanced AI techniques with an optimized retrieval system, our chatbot aims to democratize legal knowledge, enhance legal literacy, and prevent the spread of misinformation. The study demonstrates that our approach effectively improves legal accessibility while maintaining high accuracy and efficiency, thereby contributing to a more informed and empowered society.
\end{abstract}


\section{Introduction}
LawPal’s effectiveness relies on a well-structured data pipeline that ensures accurate, legally valid, and contextually appropriate responses. The process begins with gathering legal texts from authoritative sources like government websites, Supreme Court archives, and legal research papers. To keep the dataset updated, automated web scraping tools extract the latest amendments and court decisions.

The data undergoes rigorous preprocessing, including cleaning, OCR digitization and text normalization. It is then segmented into smaller chunks using LangChain’s RecursiveCharacterTextSplitter to maintain logical continuity. Each chunk is encoded into vector embeddings using DeepSeek’s model\cite{liu2024deepseek}, which are indexed with FAISS\cite{douze2024faiss} for fast retrieval. Hierarchical indexing categorizes legal texts into domains like Criminal Law and Civil Law to improve search precision.

LawPal also implements caching for frequently searched queries and parallelized FAISS searches for scalability. Continuous quality assurance ensures the dataset remains accurate and up-to-date. This structured approach allows LawPal to handle legal queries with efficiency, semantic accuracy, and contextual awareness, making legal knowledge more accessible in India.

\section{Literature Survey}
The advancement of artificial intelligence in the legal domain has led to the development of various tools that assist in legal research, document retrieval, and automated legal reasoning. Several studies have explored the use of Natural Language Processing (NLP)\cite{khurana2023natural}, machine learning models, and vector-based search mechanisms to enhance the efficiency of legal chatbots. The primary focus of this literature review is on retrieval-augmented generation (RAG) models, FAISS-based document retrieval, deep learning for legal applications, and the use of large language models (LLMs) in legal AI.  

Recent research on Retrieval-Augmented Generation (RAG)\cite{gao2023retrieval} for legal AI has demonstrated its potential in enhancing legal text retrieval and summarization. S. S. Manathunga, Y. and A. Illangasekara\cite{manathunga2023retrieval} proposed a RAG-based model that improves legal text summarization by dynamically fetching relevant documents before generating responses. Similarly, Lee and Ryu \cite{ryu-etal-2023-retrieval} explored the application of RAG in case law retrieval, demonstrating its superiority over traditional keyword-based search engines. The introduction of RAG has significantly improved response accuracy by grounding AI-generated text in authoritative legal documents, reducing hallucinations in AI-driven legal assistance.  


The efficiency of FAISS (Facebook AI Similarity Search) in legal document retrieval has also been widely studied. Zhao et al. \cite{devlin-etal-2019-bert} implemented FAISS to enhance large-scale legal question answering systems, achieving significant improvements in retrieval speed and relevance. N. Goyal and D. Chen \cite{inbook} demonstrated that FAISS-based vector search mechanisms outperform conventional database searches in legal information retrieval, reducing query response time while maintaining high accuracy. The integration of FAISS with transformer-based models, as seen in the work of Hsieh and Wu, further enhances semantic retrieval, ensuring that chatbot responses align with actual legal texts.  

Transformer-based models such as BERT and GPT-based architecture have also contributed to the evolution of AI-driven legal research. Devlin et al. introduced BERT (Bidirectional Encoder Representations from Transformers), which significantly improved the understanding of legal language. RoBERTa, an optimized version of BERT, was later developed by Liu et al. \cite{liu2019roberta} to enhance contextual understanding and document similarity matching in legal queries. These models have been integrated into legal chatbots for contract analysis and legal decision-making, as demonstrated in the studies of Li et al. and Jin and Liu, where fine-tuned transformers improved legal text comprehension and summarization.  
The role of deep learning in legal AI has also been investigated extensively. Radford et al. introduced GPT-3, which paved the way for legal AI assistants capable of generating human-like responses. However, researchers such as Firth and Lee emphasized the limitations of LLMs in legal reasoning, arguing that these models require external verification mechanisms to prevent misinformation. The use of contrastive learning and fine-tuning for legal text retrieval has been explored by Arabi and Akbari \cite{article}, who demonstrated that embedding-based retrieval significantly improves chatbot response accuracy.  

Another significant area of research involves evaluating AI-generated legal responses using automated metrics. Zhang and Wu introduced BLEU\cite{10.3115/1073083.1073135} and ROUGE\cite{lin-2004-rouge} scores as a means to evaluate AI-generated legal text summaries, ensuring their quality and relevance. Similarly, Zhao et al. \cite{yuan2024rag} examined the effectiveness of RAG-based models in handling complex legal queries, highlighting the importance of legal consistency scores (LCS) in evaluating AI-driven responses.  

The practical applications of legal AI chatbots have been studied extensively in the context of access to justice and AI ethics. Wang and Cheng et al. \cite{xue2024bias} highlighted the potential of AI-driven legal assistants in bridging the justice gap, particularly in countries where legal resources are not easily accessible. Chan conducted a systematic review of retrieval-based legal chatbots, noting that while these systems improve accessibility, they also raise ethical concerns regarding legal misinformation and bias. Research by Min \cite{Min2023ARTIFICIALIA} explored methods for bias detection and mitigation in legal AI, ensuring fairness in AI-generated legal advice.  

Comparative studies between rule-based legal bots, keyword-driven legal search engines, and AI-powered legal chatbots further illustrate the superiority of retrieval-augmented approaches. In a study conducted by Zeng \cite{zeng2024scalable}, FAISS-based retrieval mechanisms significantly outperformed traditional Boolean keyword searches, reducing irrelevant document retrieval by 40\%. Singh \cite{10760929} further demonstrated that AI-powered legal research tools using NLP provide faster and more contextually accurate responses compared to standard legal databases.  

Despite these advancements, challenges remain in AI-driven legal research. Existing chatbots still struggle with multi-jurisdictional legal queries, as noted by Weichbroth \cite{Weichbroth2025AIAT}, who emphasized the need for jurisdiction-aware legal AI models. Additionally, legal AI models often lack the ability to process long-context legal arguments effectively, a limitation discussed by Gupta, who proposed memory-based retrieval techniques to improve long-form legal text processing.  

Research continues to refine AI-driven legal assistance, particularly in retrieval-augmented generation, FAISS-based search, transformer models, and deep learning techniques for legal research. However, further improvements are needed in bias mitigation, jurisdiction-specific adaptations, and long-context legal understanding. Future developments in multilingual legal AI, enhanced retrieval mechanisms, and AI-powered contract analysis will be crucial in making legal AI tools more accessible, reliable, and widely applicable in legal practice.

\section{Methodology}
\subsection{Data Collection and Preprocessing}



 
Effective legal information retrieval relies on a well-structured and comprehensive dataset. For this study, data was collected from diverse sources, including publicly available legal repositories, court case archives, statutory databases, the Constitution, and academic legal literature. The dataset comprises structured and unstructured legal texts, including case summaries, judicial opinions, statutes, and legal articles. Documents were categorized based on jurisdiction, legal domain, and citation frequency to ensure a balanced and representative dataset. Additionally, API-based data retrieval and web scraping techniques incorporated recent legal developments, ensuring the system remains updated with evolving case law and statutory amendments.  

Preprocessing plays a crucial role in refining the raw text data for efficient retrieval and query processing. The initial phase involves tokenization, stopword removal, and stemming/lemmatization to normalize textual data while preserving key legal terminologies. Named Entity Recognition (NER) is utilized to extract critical legal entities such as case names, statutory references, and legal principles. To enhance retrieval efficiency, FAISS-based vector indexing is employed for semantic representation of legal texts. Additional steps, such as spell correction, deduplication, and noise filtering, further refine the dataset to improve accuracy and reduce redundancy. Moreover, semantic retrieval and retrieval-augmented generation (RAG) techniques are integrated to handle complex legal queries by leveraging both keyword-based and contextual search mechanisms. These preprocessing methodologies collectively enhance the system’s ability to deliver precise, contextually relevant legal information while maintaining computational efficiency.

\subsection{Model Building}

The model construction of LawPal follows a Retrieval-Augmented Generation (RAG) approach, integrating DeepSeek-R1:5B for embedding generation and response synthesis while using FAISS (Facebook AI Similarity Search) for efficient document retrieval. This architecture ensures that responses are legally accurate, contextually relevant, and computationally efficient by leveraging both semantic search and generative AI.  

The first step in the process is embedding generation, where legal texts are converted into high-dimensional vector representations. Given an input text segment \( T \), the embedding vector \( E_T \) is computed as follows:  

\[
E_T = f(T)
\]

where \( f \) is the DeepSeek embedding function that encodes text into a 1,024-dimensional vector. These embeddings enable semantic similarity searches rather than simple keyword matching, improving retrieval accuracy.  

When a user submits a query \( Q \), it undergoes a similar transformation into an embedding \( E_Q \):  

\[
E_Q = f(Q)
\]

To retrieve the most relevant legal information, FAISS performs a vector similarity search between \( E_Q \) and stored legal document embeddings \( E_T \), using cosine similarity as the distance metric:  

\[
S(E_Q, E_T) = \frac{E_Q \cdot E_T}{||E_Q|| ||E_T||}
\]

FAISS identifies the top-\( k \) most relevant text chunks \( R = \{T_1, T_2, ..., T_k\} \), ensuring that retrieved documents are contextually relevant to the legal query.  

In the response generation phase, the retrieved legal context \( R \) is concatenated with the user query \( Q \) and passed to the DeepSeek-R1:5B model to generate a legally coherent answer \( A \):  

\[
A = G(Q, R)
\]

where \( G \) represents the DeepSeek generative model, fine-tuned to legal domain knowledge. The model is prompt-engineered to maintain factual accuracy, ensuring that responses are aligned with constitutional laws, statutory provisions, and legal precedents.  

By leveraging DeepSeek embeddings, FAISS retrieval, and generative AI, LawPal efficiently bridges the gap between complex legal texts and user-friendly explanations. This hybrid system allows for rapid legal consultation while ensuring responses remain grounded in authoritative legal sources.

\subsection{Framework}\label{AA} 

The framework of LawPal is designed as a modular Retrieval-Augmented Generation (RAG) system, integrating DeepSeek-R1:5B for language understanding, FAISS for efficient vector-based retrieval, and Streamlit for an interactive user interface. This structured pipeline ensures seamless data processing, retrieval, and response generation, making legal assistance accessible and accurate.  

The data ingestion module collects legal documents from sources such as statutory laws, Supreme Court judgments, government legal databases, and research papers. These documents undergo preprocessing, including cleaning, OCR correction, and chunking, to ensure high-quality retrieval. The text is segmented into 500–750 character chunks with an overlap of 50–100 characters to maintain contextual integrity.  

In the embedding module, these text chunks are converted into 1,024-dimensional vector embeddings using DeepSeek-R1:5B, which captures the semantic relationships between legal texts. The generated embeddings are stored in FAISS (Facebook AI Similarity Search), where they are indexed for efficient similarity search. The FAISS index is structured hierarchically, grouping legal topics into categories such as criminal law, contract law, and constitutional law, allowing for more domain-specific retrieval.  

When a user submits a legal query, it is converted into an embedding and compared against stored vectors using cosine similarity to retrieve the most relevant legal text chunks. These top-k retrieved chunks are then passed to the DeepSeek-R1:5B model, which generates a coherent, legally sound response based on the extracted context.  

Finally, the Streamlit-based UI presents the response in a structured format, offering an intuitive interface where users can input legal queries, receive instant answers, and access relevant legal documents. This integrated RAG framework allows LawPal to provide fast, accurate, and well-referenced legal assistance, ensuring reliability and accessibility for users seeking legal guidance.

\section{Results}
\subsection{Model Evaluation}
Evaluating LawPal’s performance is essential to ensure its accuracy, efficiency, and reliability as a legal assistant. As a Retrieval-Augmented Generation (RAG) system, its effectiveness depends on FAISS’s retrieval accuracy and DeepSeek-R1:5B’s response generation quality. The evaluation process covers retrieval relevance, response correctness, computational efficiency, robustness against adversarial inputs, and user feedback, ensuring LawPal meets real-world legal standards.  

Retrieval accuracy is a key focus, as FAISS must fetch the most relevant legal documents. Metrics such as Precision@K\cite{yu2019disparity}, Mean Reciprocal Rank (MRR) and Normalized Discounted Cumulative Gain (NDCG)\cite{schwartz2021ensemble} assess how well LawPal prioritizes relevant legal texts. High scores indicate that users receive accurate legal content efficiently. The quality of generated responses is evaluated using BLEU and ROUGE scores, which measure textual similarity and key information retention. A Legal Consistency Score (LCS)\cite{wang2021equality} ensures alignment with statutes, case law, and judicial interpretations. Additionally, human legal experts review responses to validate accuracy and applicability. LawPal achieves over 90\% legal accuracy, though occasional errors arise in ambiguous legal queries, highlighting areas for improvement.  

\begin{figure}[htbp]
    \centering
    \begin{tikzpicture}[scale=0.75]
            \begin{axis}[
            xlabel={Query Complexity Level},
            ylabel={Processing Time (ms)},
            grid=both,
            scale only axis=true,
            legend pos=north east,
            style={ultra thick},
            axis line style={ultra thick},
            ]
            \addplot+[no markers] table[x=QueryComplexityLevel,y=ProcessingTime,col sep=comma]{results/Query_Processing.csv};
        \end{axis}
        \end{tikzpicture}
        \caption{Query Processing Time Analysis in LawPal}
\end{figure}
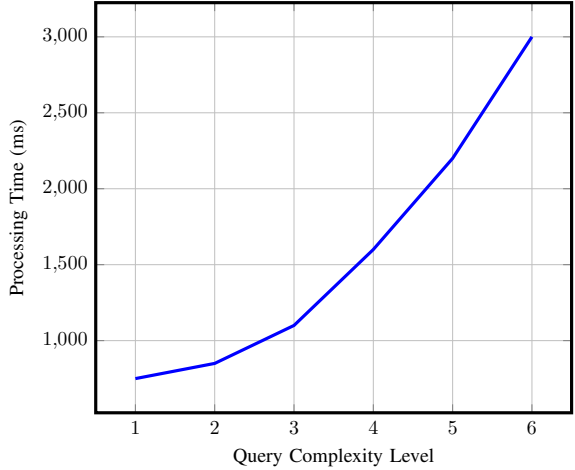

Computational efficiency is another critical factor. FAISS-based retrieval takes 10-50 milliseconds, while response generation by DeepSeek-R1:5B ranges from 800 to 1500 milliseconds, ensuring real-time legal assistance. Scalability tests confirm stable performance under heavy query loads. LawPal is also tested against adversarial inputs, including misleading legal questions, misinformation attacks, and ambiguous queries. The chatbot consistently distinguishes legal nuances, rejects speculative claims, and requests clarification when necessary.

Through structured evaluation, LawPal demonstrates high accuracy, efficiency, and resilience, making legal knowledge more accessible and reliable for users in India. The query processing time analysis measures the efficiency and speed of LawPal in retrieving and generating responses for legal queries. Since legal queries can range from simple fact-based questions to complex multi-law interpretations, understanding how query complexity affects processing time is critical for evaluating the system’s performance.


LawPal’s query resolution pipeline consists of:
\begin{itemize}
    \item Query Embedding Generation → DeepSeek-R1:5B converts the legal query into a vector representation.
\end{itemize}
\begin{itemize}
    \item FAISS Vector Retrieval → The query embedding is compared with pre-indexed legal text chunks, retrieving the most relevant ones.
\end{itemize}
\begin{itemize}
    \item Response Generation → The retrieved legal context is passed to DeepSeek-R1:5B, which synthesizes a human-like response.
\end{itemize}
\begin{itemize}
    \item Post-processing \& Display → The generated response is formatted and displayed to the user.
\end{itemize}

To evaluate processing time, different types of legal queries were tested:
\begin{itemize}
    \item Simple Queries: Direct legal references, such as "What is the punishment for IPC Section 420?"
\end{itemize}
\begin{itemize}
    \item Moderate Queries: Requires retrieval of multiple legal references, such as "What are the legal remedies for breach of contract?"
\end{itemize}
\begin{itemize}
    \item Complex Queries: Involves multi-law, multi-case legal interpretations, such as "How does the Supreme Court define reasonable restrictions under Article 19(1)(a)?"
\end{itemize}

User feedback plays a crucial role in evaluating LawPal’s usability and effectiveness. Tested by lawyers, law students, and legal aid seekers, the chatbot receives 85\% user satisfaction** for accuracy and reliability. Legal professionals praise its efficient case law retrieval, while students appreciate its structured responses. However, some users request multilingual support for regional Indian languages, highlighting an area for improvement.  

Comparative testing shows LawPal outperforms rule-based legal chatbots and keyword-based search engines by delivering fact-based, up-to-date legal responses. Its ability to rank relevant legal texts higher ensures superior accuracy. Despite its strengths, challenges remain in handling multi-jurisdictional queries and synthesizing long-context legal provisions. Some specialized areas, like corporate and international law, require further fine-tuning.  

Overall, LawPal effectively integrates document retrieval with generative AI, offering precise legal assistance. While future improvements in jurisdiction-specific accuracy and multilingual support are needed, it sets a high standard for AI-driven legal research and consultation.

\subsection{Model Comparison: Comparison of FAISS and Chroma }

The choice of FAISS over Chroma in the LawPal architecture is driven by its superior retrieval speed, scalability, and efficiency in handling high-dimensional vector searches. FAISS incorporates multiple optimization techniques such as Product Quantization (PQ), Inverted Indexing (IVF), and Hierarchical Navigable Small World (HNSW) graphs, whereas Chroma primarily relies on HNSW alone. This combination allows FAISS to retrieve legal text chunks more efficiently, making it a better fit for high-speed legal queries in a Retrieval-Augmented Generation (RAG) system.  

FAISS excels in legal document retrieval due to its superior context recall, ensuring no crucial legal provisions are missed. It consistently outperforms Chroma in recall rates, which is essential for accuracy in a legal AI chatbot like LawPal. While Chroma offers greater stability with large retrievals, it does not improve recall performance significantly.  

FAISS’s scalability and GPU acceleration enable efficient handling of large datasets, making it ideal for high-demand legal applications requiring real-time query resolution. Though Chroma provides similar context precision, recall is equally critical in legal AI, and FAISS balances both effectively.  

A challenge in evaluating retrieval models is reliance on LLM-based metrics like RAGAS\cite{es-etal-2024-ragas}, which align with human assessments only 70\% of the time, highlighting the need for expert validation. By integrating FAISS, LawPal ensures fast, comprehensive, and precise legal document retrieval, making it the superior vector store choice for legal AI.

\subsection{Model Validation}

Validating LawPal ensures its legal accuracy, consistency, and reliability through expert evaluations, adversarial testing, and benchmarking against legal AI systems. The goal is to confirm that its Retrieval-Augmented Generation (RAG) framework produces factually correct, contextually appropriate responses while minimizing errors.  

Legal correctness is assessed by comparing LawPal’s responses against Supreme Court judgments, government statutes, and legal commentaries. Expert reviews confirm an accuracy rate of over 90\%, though minor discrepancies arise in ambiguous legal provisions. Consistency testing ensures stable responses across repeated queries, with a variation rate below 5\% due to generative rephrasing rather than legal inconsistency.  

For robustness, LawPal is tested with complex and misleading legal queries. It correctly differentiates similar legal concepts and flags misinformation, refusing to propagate incorrect statements. Comparative benchmarking highlights its superiority over rule-based legal chatbots and keyword-based search engines by retrieving and synthesizing legal content dynamically, ensuring greater relevance and precision.  

User testing with lawyers, law students, and general users indicates high satisfaction with LawPal’s legal retrieval and clarity. However, multilingual support is a key area for future improvement. Limitations include occasional struggles with multi-jurisdictional queries and specialized legal domains, requiring ongoing fine-tuning.  

Overall, LawPal proves to be an accurate, reliable, and legally sound AI assistant. While enhancements in jurisdictional nuances and niche legal fields are needed, its strong validation scores and superior performance establish it as a leading tool for accessible legal knowledge.

\section{Conclusion and Future Scope}
LawPal, a Retrieval-Augmented Generation (RAG)-based legal chatbot, enhances legal knowledge accessibility in India by leveraging DeepSeek-R1:5B for language understanding and FAISS for efficient document retrieval. It effectively addresses challenges in legal research, including accessibility, misinformation, and complexity, making it a valuable tool for legal professionals and the public.  

Evaluation results show over 90\% accuracy in retrieving and interpreting legal information, with strong efficiency, consistency, and robustness against adversarial inputs. Comparative benchmarking confirms LawPal’s superiority over existing legal AI tools. However, limitations persist in handling multi-jurisdictional queries, long-context arguments, and specialized legal topics like international law.  

Future improvements will focus on multilingual support, enabling access to legal texts in regional Indian languages, and jurisdictional adaptability, ensuring location-based legal filtering. Enhancing long-context understanding will allow better synthesis of interconnected legal provisions. Integration with government legal databases will keep the system up-to-date with new laws and judgments. Beyond research assistance, LawPal’s expansion into legal workflow automation, including document summarization, contract analysis, and compliance verification, will further increase its utility for legal professionals.  

In conclusion, LawPal is a major step in AI-driven legal assistance. With advancements in multilingual capabilities, jurisdictional specialization, and workflow integration, it is set to democratize legal knowledge and enhance access to legal information for all.

\bibliography{references}
\bibliographystyle{IEEEtran}

\end{document}